# IMAGE FUSION TECHNOLOGIES IN COMMERCIAL REMOTE SENSING PACKAGES

Firouz Abdullah Al-Wassai[1] and N.V. Kalyankar[2]

[1]Research Student, Computer Science Dept., (SRTMU), Nanded, India
[2]Principal, Yeshwant Mahavidyala College, Nanded, India
fairozwaseai@yahoo.com , drkalyankarnv@yahoo.com

**Abstract:** Several remote sensing software packages are used to the explicit purpose of analyzing and visualizing remotely sensed data, with the developing of remote sensing sensor technologies from last ten years. According to literature, the remote sensing is still the lack of software tools for effective information extraction from remote sensing data. So, this paper provides a state-of-art of multi-sensor image fusion technologies as well as review on the quality evaluation of the single image or fused images in the commercial remote sensing packages. It also introduces program (ALwassaiProcess) developed for image fusion and classification.

## INTRODUCTION

Automatic recognition, description, classification, and grouping of patterns are important problems in a variety of engineering and scientific disciplines such as statistics, computer-aided diagnosis, marketing, computer vision, bio-medicine, and remote sensing. This topic has been extensively studied and applied to several tasks in various areas, with continual evolution of computers and sensors, it has become increasingly important to understand the interactions and associations between data from different sensors and deduce meaningful inferences. Remote sensors offer a wide variety of image data with different characteristics in terms of temporal, spatial, radiometric and spectral resolutions. Very high spatial resolution enables an accurate description of shapes, features and structures while high spectral resolution enables better identification and discrimination of the features based on their spectral response in each of the narrow bands. A current remote sensor offers multi-spectral sensors and advanced multi-spectral sensors called hyper-spectral sensors, it detects hundreds of very narrow spectral bands throughout the visible, near-infrared, and mid-infrared portions of the electromagnetic spectrum. also, high spatial ranges from (0.5m-10m) in remote sensing commercial domain such as IKONOS, QuickBird, OrbView-3 and SPOT-5…etc. despite that more than 80% of the modern earth observation satellite sensors and many airborne digital cameras simultaneously, offers a tradeoffs between high spatial and high spectral resolution, and no single system offers both (i.e. collects high-resolution panchromatic (Pan) and low-resolution multispectral (MS) images or the opposite). For example TM Sensors on board the Landsat-series of satellites imagery has significant advantage in 6 spectral wavebands but is very poor in spatial resolution (30m) for certain applications, whereas The HRV sensors on board the SPOT-series of satellites performed very well in spatial resolution 10m but with low-spectral resolution, as a single-channel PAN image and three multispectral images at spatial resolution 20m. In order to automate the processing of these satellite images the concepts for image fusion are needed.

The term “image fusion” covers multiple techniques used to combine the geometric detail of a high-resolution panchromatic image and the colour information of a low-resolution multispectral image to produce a final image with the highest possible spatial information content, while still preserving good spectral information quality. Multi-sensor image fusion is widely recognized as an efficient tool for improving overall performance in image based application. Some of the common applications of image fusion includes; visual display enhancement, pattern recognition, machine learning, Object separation[1], texture information[2] image categorization and retrieval [3], Classification of Forest [4], and vehicle detection [5.], targets tracking [6], ….etc. The applications themselves usually used to drive the choice of associated with fusion algorithms. Although advancements in image fusion have been made, the research has established that no single fusion algorithm can work for every application; because the sensors of images themselves have varying responses under different operating conditions such as merging images of multi-spectral differ with merging hyper-spectral satellite sensor data. As a result, a lot of variations of image fusion techniques appeared in the lecturer. Thus, associated algorithms are generally tuned toward specific tasks and leading to custom solutions as well as depends upon the user’s experience. Image fusion is only an introductory stage to another task Therefore; the performance of the fusion algorithm must be measured in terms of improvement or image quality to evaluate the possible benefits of fusion. With increasingly sophisticated algorithms, the need for a cost-effective and standardized on designing software for image Fusion and quality evaluate the possible benefits of fusion.

Recently, quite a few survey papers have been published, providing overviews of the history, developments, Evaluation, and the current state of the art of image fusion in the image-based application fields [7-11]. But recent development of multi-sensor data fusion in remote sensing software packages has not been discussed in detail. The objectives of this paper are to pre-sent an overview of the image fusion tools in the software packages are available for the explicit purpose for visualizing of remote sensing images. Focuses mostly on image fusion and quality evaluation of the single image or fused images components of these packages. The subsequent sections of this paper are organized as follows: section II gives the brief overview of the textbooks with their software implementation. III covers the review on Image Processing Systems; IV This section gives a quick overview the program ALwassaiProcess for image fusion and classification and is subsequently followed by the conclusion.

## REVIEW ON THE TEXTBOOKS WITH THEIR SOFTWARE IMPLEMENTATION

There are many textbooks in image processing which include a software implementation include: Lindley, C. A.: Practical Image Processing in C [12] and Pitas, I.: Digital Image Processing Algorithms [13] which both cover basic image processing and computer vision algorithms. Parker, J. R.: Practical Computer Vision Using C [14] offers an excellent description and implementation of low-level image processing tasks within a well-developed framework, but again does not extend to some of the more recent and higher level processes in computer vision in his later text Image Processing and Computer Vision such as image [15]. In Computer Vision and Image Processing [16] takes an applications-oriented approach to computer vision and image processing, offering a variety of techniques in an engineering format, together with a working package with a GUI. However, image fusion techniques were absent. Image Processing in Java [17], concentrates on Java only and concentrates more on image processing systems implementation than on feature extraction (giving basic methods only) as the previous the image fusion was absent. Masters, T.: Signal and Image Processing with Neural Networks– a C++ Sourcebook [18] offers good guidance in combining image processing technique with neural networks and gives code for basic image processing technique, such as frequency domain transformation. The newest textbook in depth of image processing of remote sensing Mather, P. M.: Computer Processing of Remotely-Sensed Images [19] offers an excellent description and implementation of image processing for Remotely-Sensed Images tasks.

Through, A hundreds of variations of image fusion techniques Publications in the lecturer it were absent *in these* previous books with their software implementation as well as the image quality to evaluate the possible benefits of fusion, except [19] has some methods as the image fusion techniques.

## REVIEW ON COMMERCIAL REMOTE SENSING PACKAGES

In recent years, a number of commercial remote sensing packages have been greatly developed, their performance and functionality owing to the advances in computing technology in direct response to the need for efficiently processing a huge quantity of remote sensing data and many are screen-based using a Windows system. There are so many Image fusion techniques among them some are used for image fusion in these system includes, intensity-hue-saturation (IHS), high-pass filtering (HPF), principal component analysis (PCA), different arithmetic combination e.g., Brovey transform (BT) and wavelet transform [20-25]. A number of mature image processing systems for illustration and description remotely sensed data, but are not limited to, include:

1. **IDRISI** is a sophisticated desktop raster geographic information and image processing system; it is developed by the Graduate School of Geography at Clark University, Worcester, and Massachusetts. The latest release, called IDRISI Andes (version 15), is 32-bit Windows NT– compatible. This affordable system comprises over 250 modules or stand-alone programs for the digital analysis and visualization of spatial data, including the remotely sensed imagery, in a single package [26]. It is widely being used by government agencies, schools, research institutions, academia, and the private sector testifies to its popularity [27]. However, IDRISI offered limited capabilities of image fusion technologies as maintained above, as well as the image quality evaluation, there was absent.

2. **ERDAS** Imagine is one of the oldest and leading geoinformatic software companies. Its major product, Imagine, contains a suite of comprehensive and sophisticated tools for digital analysis of remotely sensed data. ERDAS Imagine is offered at three levels, Essentials, Advantage, and Professional. Imagine Essentials encompasses a set of powerful tools for manipulating geographic and imagery data, such as image georeferencing, visualization, and map output. Imagine Advantage extends the capabilities of Imagine Essentials through addition of several more functions, such as mosaicking, surface interpolation, advanced image interpretation. Imagine Professional contains more classification and spectral analysis, and radar processing utilities than Imagine Essential. In addition to all those modules included in both Essentials and Advantage, it also encompasses add-on tools for complex image analysis and modeling, radar data analysis and advanced classification [26]. Despite that ERDAS Imagine has the limited range of image fusion it offers the most popular methods as the previous system, and the image quality evaluation for evaluation the possible benefit of fusion it was absent. Also, with many add-on modules, ERDAS image could be quite expensive.

3. **ENVI**: Produced by Research Systems (now called ITT Visual Information Solutions), ENVI is a comprehensive icon-driven image analysis package designed especially for processing large multispectral and hyperspectral remote sensing data [28]. ENVI contains in image processing functions for image fusion, but with the most popular methods as main-tained above. As an important image analysis system, ENVI offers a wide range of traditional and nontraditional image classifiers. For instance, it is one of the earliest systems with the capability of image classification based on machine learning. At present ENVI is able to classify images using neural network and binary encoding [26]. However, image quality evaluation for evaluation the possible benefit of fusion it was absent. Also, the operation of ENVI can be further improved via better organization of some functions such as image display; Graphic icons may be added to accompany the functions in the toolbar for quick access.

4. **ER Mapper** is an Australian software company specializing in digital image analysis. Its major product of the same name is a powerful window-based imagery analysis package that offers a complete suite of image processing tools [29]. However, ER Mapper has a limited range of image fusion functions. ER Mapper offer the evaluated of classified images results for their accuracy. The produced confusion matrix shows the producer's and user's accuracy, as well as the overall accuracy. But no special modules are available for assessment of the other images quality.

5. **PCI Geomatica**: Headquartered in Toronto, Canada, PCI is a leading software developer specializing in remote sensing, digital photogrammetry, and cartography. This package of an extensive suite of geospatial tools offers the most complete geospatial solution via its many built-in capabilities. The strength of PCI lies in its high degree of automation and customized workflows [26]. However, this advantage turns to a limitation at the same time as it can be applied to a narrow range of image fusion and the assessment of the image quality it was absent  as the previous systems.

6. **GRASS GIS**: commonly referred to as GRASS (Geographic Resources Analysis Support System), is free Geographic Information System (GIS) software used for geospatial data management and analysis, image processing, graphics/maps production, spatial modeling, and visualization. GRASS GIS is currently used in academic and commercial settings around the world, as well as by many governmental agencies and environmental consulting companies. GRASS GIS is an official project of the Open Source Geospatial Foundation (OSGeo) [30]. Also has a limitation of image fusion technique as well as the assessment of the image quality it been absent.

## SUMMARY OF PROGRAM ALWASSAI-PROCESS FUNCTIONS

This program has been created to serve as a laboratory to implement routines of image processing. ALwassaiProcess runs in Window and free software for academic. Its interface and implemented in VB6 Aiming simplicity at simplicity we did not use complex classes to model images and other objects, so the code looks like an extended C++ or C. In several situations ALwassaiProcess can be the tool to develop new Image Processing algorithms, articles, thesis and dissertations.

This section gives a quick overview of running ALwassaiProcess. The Starting ALwassaiProcess by Installation of ALwassaiProcess under Windows, ALwassaiProcess is available as a standard installer package. Assuming that ALwassaiProcess is installed in the PATH, you can start ALwassaiProcess by double clicking on the ALwassaiProcess application link (or shortcut) on the desktop. When starts ALwassaiProcess it appears with the main window as shown in Fig.1, It include, menu bar and Tool bar. The Toolbar includes some Icons for quick access to the functions in menu bar items.

The functions provided by ALwassaiProcess are accessed from the main menu bar items as the following: File, Edit, View, Image Processing, Filter, Transformation, Classification, Measure and analysis, Tools, Window and help.

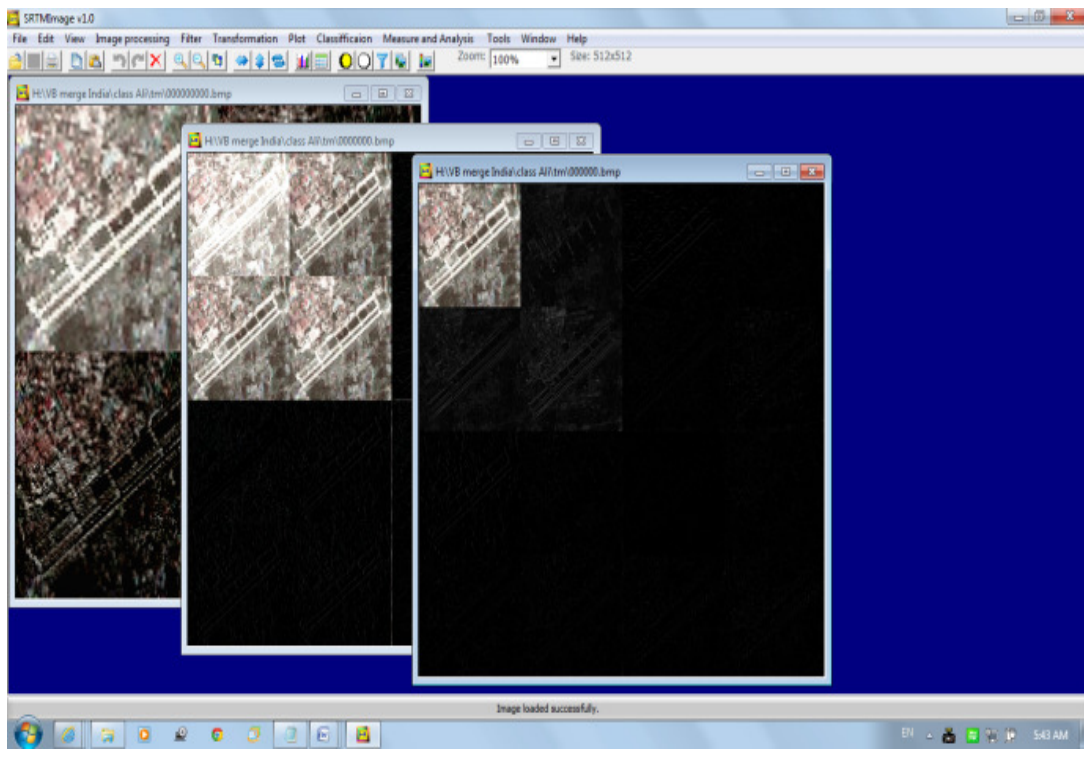

**Fig.1** The Main Window, showing the menu bar at the top, Tool bar and children window containing a Composite RGB for landsat-TM.

There are many functions provided for image processing in the program, but we cannot describe it properly, because of limitations. But here few snapshots are illustrated for some functions it's appeared in Fig.2-8.

This program exactly interested with image fusion and classification. The most important thing is introduced in it that is evaluation of images for different purpose in academic research subject.  The goal of this program provides the functions for automatic image fusion and classification, which is difficult as well as limited in the previous software as maintained. And some of them are very costly. Here, the program is cover most the effective algorithms in image fusion as well as in classification.  There is a need for image fusion or classification that are manipulations such as re-sampling radiometric correction, noise reduction, filtering,....etc, this program offers it or above things. We would like to maintain in the program, the filtering methods work corresponding to textbooks of image processing.

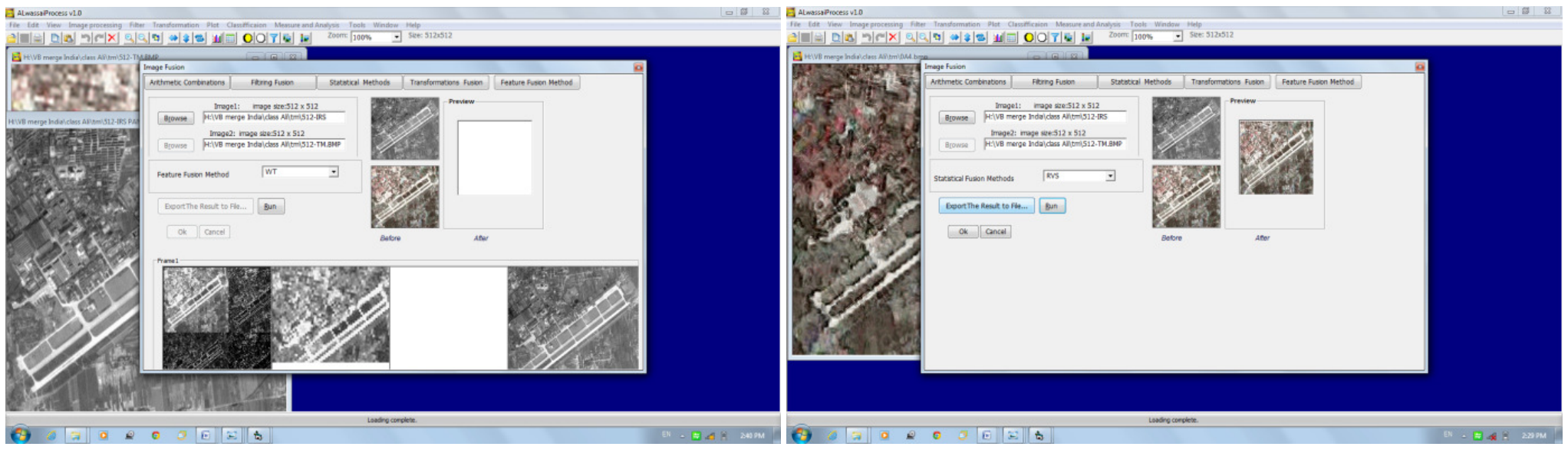

**Fig. 2:** Examples for some functions such as WV and Automatically Displays the Results of Fuse as well as Export the Results by Icon.

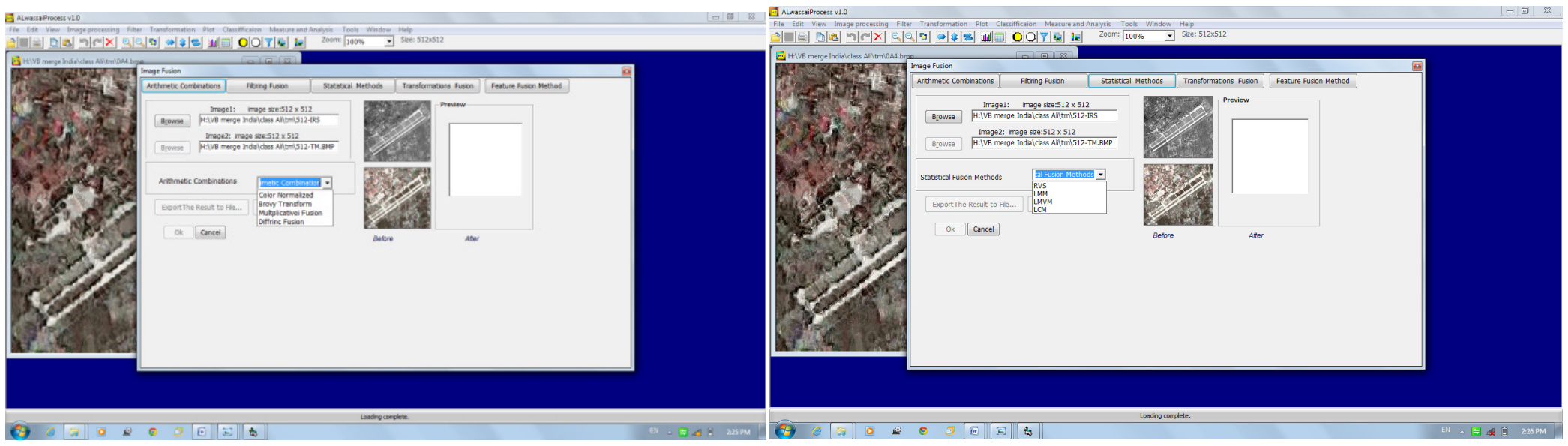

**Fig. 3** Snapshots Illustrated the Arithmetic Combination and Statistical Fusion Methods Menu Respectively.

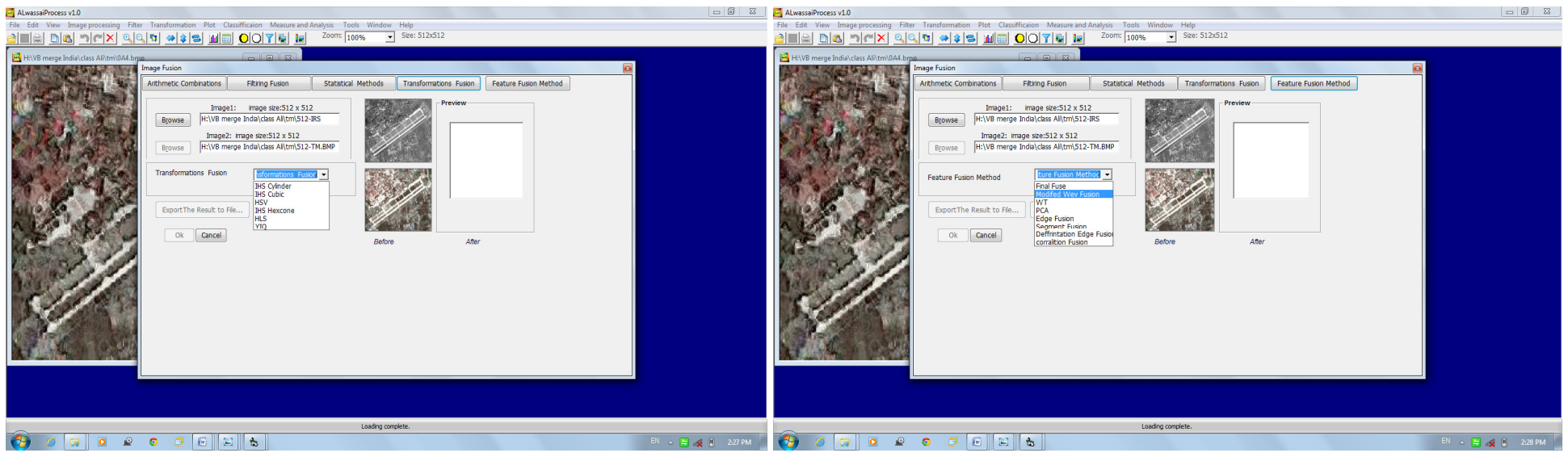

**Fig. 4** Snapshots Illustrated the Component Substitution and Feature Fusion Menu Respectively.

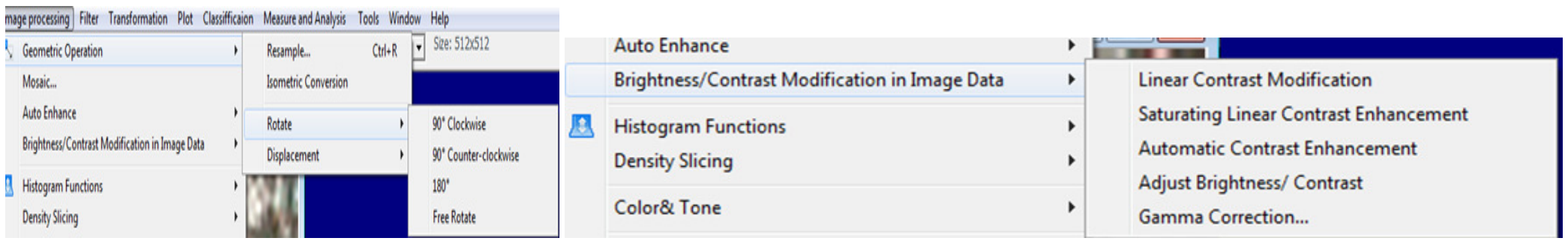

**Fig. 5a**

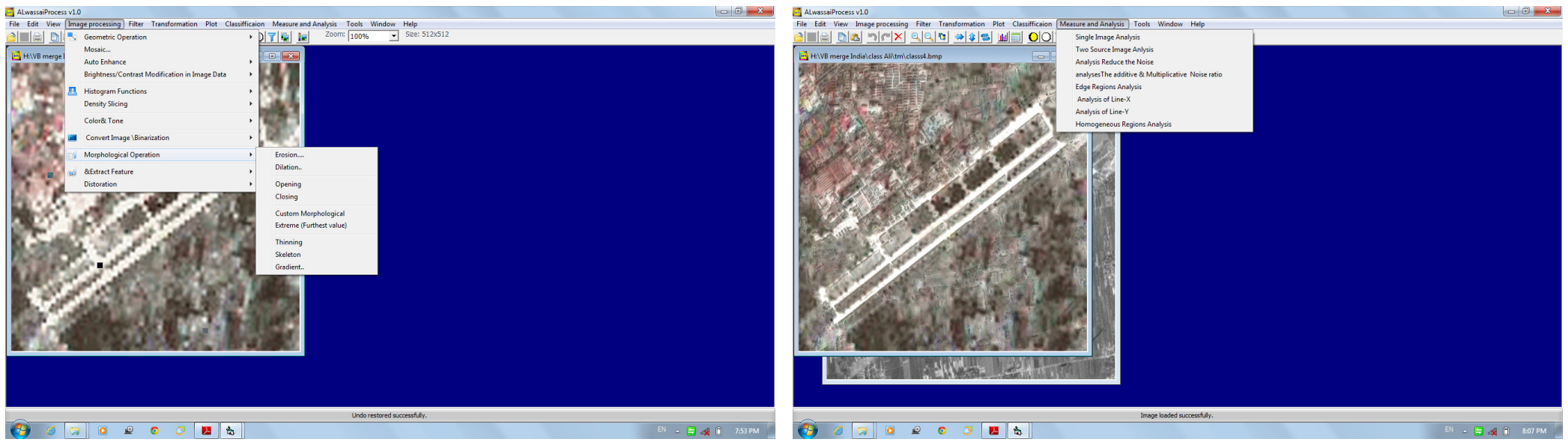

**Fig. 5b**

**Fig. 5(a, b)** Snapshots Illustrated Some Function in the Program.

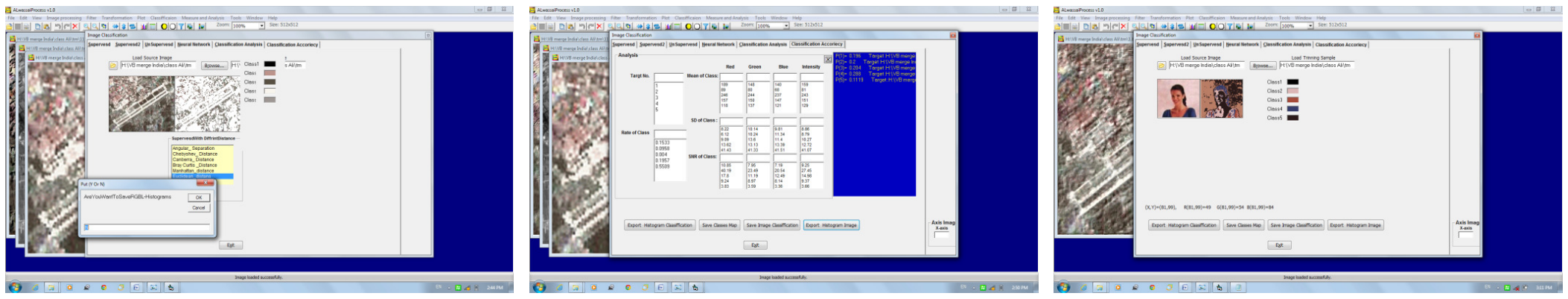

**Fig. 6** Snapshots Illustrated Some Functions for Classification.

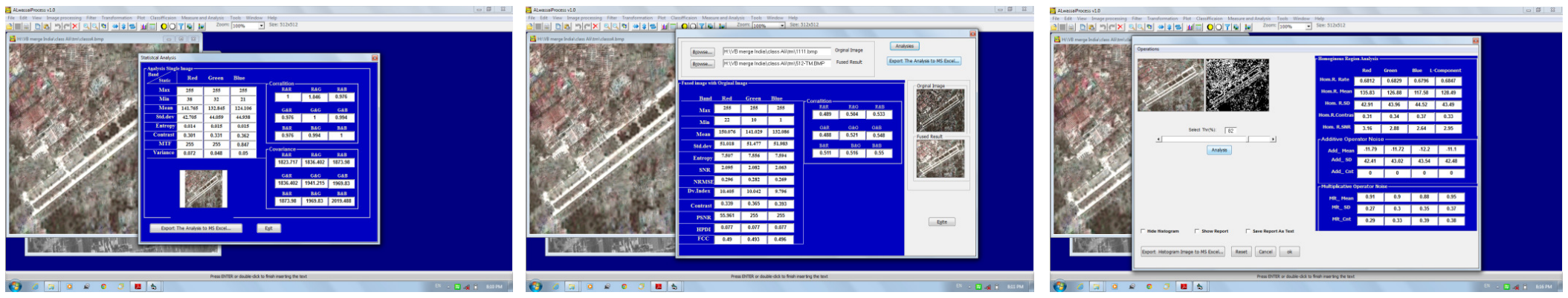

**Fig. 7a**

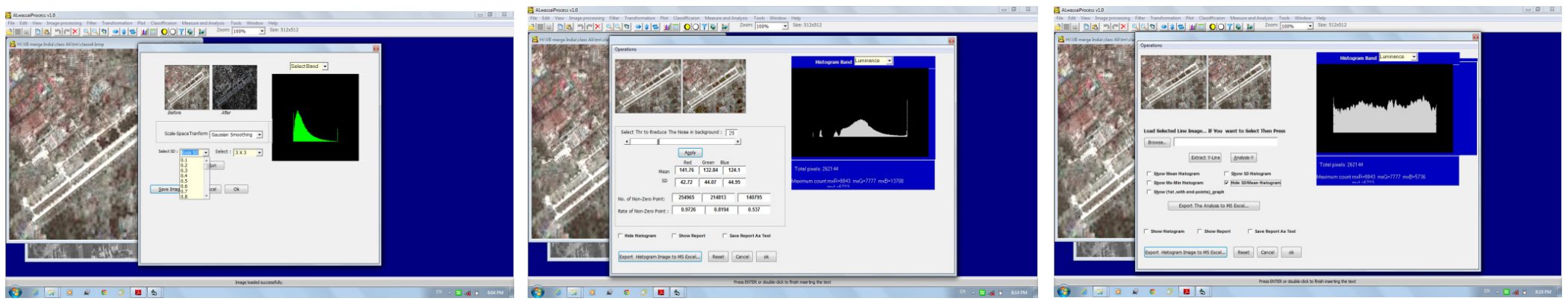

**Fig. 7b**

**Fig. 7(a, b) Snapshots** Illustrated Automatic quality assessment for Some Functions.

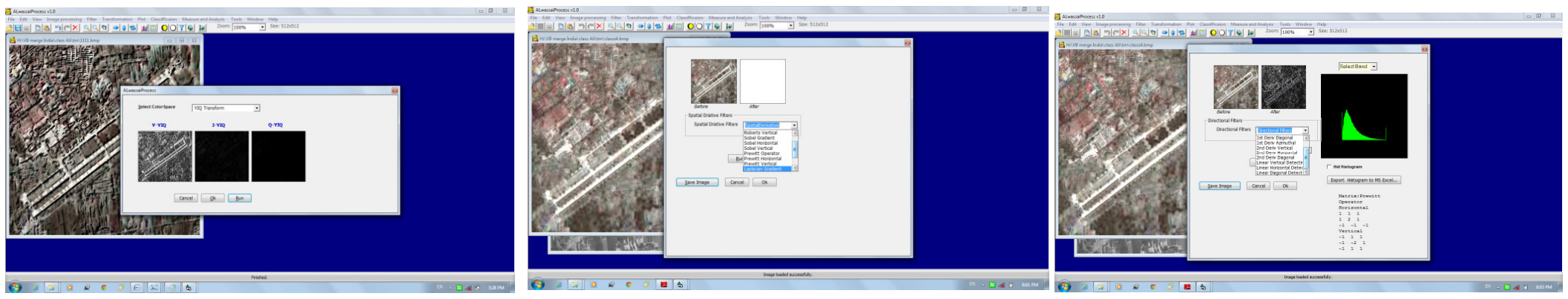

**Fig. 8a**

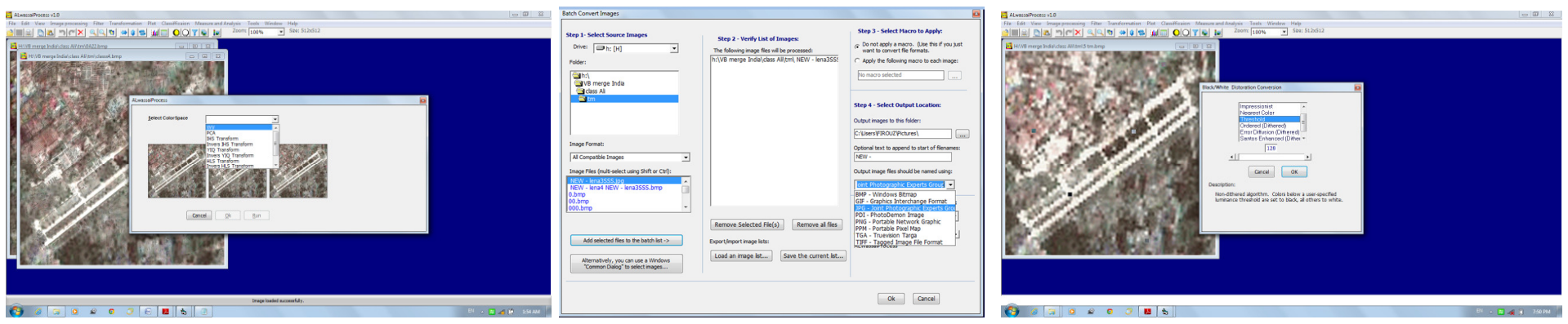

**Fig. 8b**

**Fig. 8 Snapshots** Illustrated Some Functions such as Filtering, Transformation and Batch Convert file.

## Conclusion

In this way the paper presenting the review on the some textbooks that are implementations the software and some commercial remote sensing packages which are working for image processing and analysis. The results of this review in the textbooks implementation software the image fusion was absent mostly there and the most of commercial remote sensing packages have been widely offer the popularity of fusion algorithms (i.e. PCA, IHS, HPF, BT and WV), despite a lot of variations of image fusion techniques. Are these methods in these products better performance than can be achieved by other techniques to make image fusion practitioners suitable or accurate enough to use remote sensing specialists communicated efficiently?

However, with the popularity of commercial remote sensing packages, it is easy to obtain processed remote sensing products based on fusion algorithms or Classification. But, it is noteworthy that these products may not be suitable or accurate corresponding to the conditions of image fusion to use. Where the result of fusion process is depends on its many applications, so we should to evaluate the benefits of fusion. Therefore, it is still urgent needed to make image fusion practitioners efficiently. Also, the result of review on commercial remote sensing packages for image evolution is absent there. So, we introduce our program which that it covers 15 techniques of image fusion to answer the previous ques-tion. Also, introduces maximums tools in the program which are needed in image fusion and classification as well as provide tools for automatic quality assessment implemented in [23-25, 31-33]

Finally, we hope that this program will be helpful in the future to them who want to study in this field and further investigations are necessary to develops image fusion algorithms.

## ACKNOWLEDGMENTS

Authors would like to thank everyone who contributed, directly or indirectly, either in the accomplishment or testing of this work. Thanks to Mr. Arvindkumar Gaikwad for helps, during this work.